\documentclass[conference]{IEEEtran}
\IEEEoverridecommandlockouts

\usepackage{cite}
\usepackage{amsmath,amssymb,amsfonts}
\usepackage{algorithmic}
\usepackage{graphicx}
\usepackage{textcomp}
\usepackage{xcolor}
\usepackage{float}

\def\BibTeX{{\rm B\kern-.05em{\sc i\kern-.025em b}\kern-.08em
    T\kern-.1667em\lower.7ex\hbox{E}\kern-.125emX}}

\begin{document}

\title{Extending Deep Event Visual Odometry with Sparse Point-Cloud Export}

\author{\IEEEauthorblockN{1\textsuperscript{st} Sajad Ashraf}
\IEEEauthorblockA{\textit{TU Berlin} \\
sajad.ashraf.1@campus.tu-berlin.de}
\and
\IEEEauthorblockN{2\textsuperscript{nd} Alireza Safdari}
\IEEEauthorblockA{\textit{TU Berlin} \\
alireza.safdarikhosroshahi@campus.tu-berlin.de}
}

\maketitle

\begin{abstract}
Event cameras are well suited for visual odometry under high-speed motion and challenging lighting conditions due to their low latency, high temporal resolution, and high dynamic range. Deep Event Visual Odometry (DEVO) demonstrated that monocular event-only odometry can achieve strong performance by combining sparse patch tracking, learned patch selection, recurrent correspondence refinement, and differentiable bundle adjustment. In this project, we extend DEVO with a sparse point-cloud export pipeline. Rather than modifying the core odometry formulation, our approach exposes the internal 3D structure already estimated by DEVO and converts it into an explicit point-cloud representation for visualization and further processing. In addition, we implement a practical workflow for data export, format conversion, and point-cloud cleanup. The resulting system preserves the original visual odometry pipeline while enabling sparse geometric scene output. Experiments on the BOARD SLOW sequence show that the exported sparse cloud is locally consistent with EMVS reconstructions, achieving high precision at a 5 cm threshold, while also highlighting the expected limitations in density, completeness, and sensitivity to accumulated odometry noise.
\end{abstract}


\section{Introduction}
Visual odometry (VO) is a fundamental component of autonomous robots, drones, and augmented reality systems because it enables continuous estimation of camera motion from visual input. Accurate pose tracking is especially important in scenarios involving fast motion, rapid viewpoint changes, and limited environmental structure. However, conventional frame-based cameras often suffer from motion blur and limited dynamic range, which can reduce the robustness of visual odometry pipelines \cite{gallego2022survey}.

Event cameras provide an alternative sensing modality that is particularly well suited to such conditions. Instead of recording full image frames at fixed rates, they asynchronously measure per-pixel brightness changes with very high temporal resolution and low latency. This sensing principle offers major advantages in high-speed motion and challenging illumination, and has therefore attracted increasing interest in robotics and computer vision \cite{lichtsteiner2008dvs,gallego2022survey}. As a result, event-based methods have been studied for feature tracking, optical flow, visual odometry, SLAM, and 3D reconstruction \cite{gallego2022survey}.

Despite these advantages, monocular event-only visual odometry remains a difficult problem. Event streams are sparse, noisy, and fundamentally different from conventional image sequences, which makes direct adaptation of frame-based pipelines non-trivial. Earlier event-based VO systems therefore often relied on strong geometric assumptions or on additional sensing modalities such as inertial measurements, stereo event cameras, or conventional image frames to increase robustness \cite{rebecq2017evo,rebecq2017vio,vidal2018ultimate,zhou2021esvo,hidalgo2022eds,guan2022plevio,chen2023esvio}. While these hybrid approaches can improve accuracy, they also increase hardware complexity and partially reduce the advantages of using a single event camera.

Recent progress in learning-based VO has shown that sparse correspondence-based formulations can achieve strong performance while remaining computationally efficient. In the frame-based domain, Deep Patch Visual Odometry (DPVO) demonstrated that tracking a sparse set of image patches, refining correspondences recurrently, and optimizing poses through differentiable bundle adjustment can produce accurate monocular odometry without relying on dense optical flow \cite{teed2023dpvo}. Building on this idea, Deep Event Visual Odometry (DEVO) extends sparse patch-based odometry to the event domain by representing events as voxel grids and introducing a learned patch selection mechanism tailored to sparse event data \cite{klenk2024devo}. DEVO is, to our knowledge, the first monocular event-only system in this line of work to show strong performance across several real-world benchmarks while using only a single event camera \cite{klenk2024devo}.

Although DEVO provides accurate camera trajectories and internally estimates sparse scene structure through tracked patches and optimized depths, its original focus lies on visual odometry rather than explicit 3D output \cite{klenk2024devo}. For many robotics and visualization tasks, however, trajectory estimation alone is not sufficient. Exportable 3D structure in the form of a point cloud is highly useful for scene inspection, debugging, mapping, and downstream reconstruction. Event-based 3D reconstruction has previously been explored in dedicated methods such as EMVS, which reconstructs scene structure from events and known poses \cite{rebecq2018emvs}. This motivates the central objective of our project: to extend DEVO such that the sparse 3D structure already estimated inside the odometry pipeline can be extracted and represented as a usable point cloud.

In this work, we present an extension of DEVO that enables sparse point-cloud export from the original event-based odometry pipeline. Rather than replacing the underlying VO formulation, our approach exposes the internal geometric representation already maintained by DEVO and makes it available for further processing. In addition, we implement a practical export pipeline that includes data extraction, format conversion, and point-cloud cleanup. In this way, the proposed system preserves the original strengths of DEVO while improving its usefulness for 3D visualization and downstream scene analysis.

The main contributions of this project are as follows:
\begin{itemize}
    \item We extend DEVO to expose its internal sparse 3D representation as exportable point-cloud data.
    \item We implement a practical processing pipeline for point-cloud generation, conversion, and cleanup.
    \item We analyze the usefulness and limitations of sparse point-cloud extraction from an event-based visual odometry framework.
\end{itemize}

The remainder of this report is structured as follows. Section~II reviews the background of event-based visual odometry and the DEVO framework. Section~III describes the proposed method and the point-cloud export concept. Section~IV presents the implementation details and pipeline design. Section~V reports the experimental results. Finally, Section~VI discusses the limitations of the approach, and Section~VII concludes the report.

\section{Background and Related Work}
\subsection{Event Cameras and Event-Based Vision}
Event cameras are neuromorphic vision sensors that operate fundamentally differently from conventional frame-based cameras. Instead of recording full images at fixed time intervals, they asynchronously output events whenever the brightness at a pixel changes beyond a sensor-dependent threshold. Each event is typically represented by pixel location, timestamp, and polarity, which indicates whether the brightness increased or decreased \cite{lichtsteiner2008dvs,gallego2022survey}. Due to this sensing principle, event cameras offer several advantages for robotic perception, including microsecond-level temporal resolution, low latency, high dynamic range, and strong robustness against motion blur \cite{lichtsteiner2008dvs,gallego2022survey}. These properties make them especially attractive for high-speed motion, aggressive maneuvers, and challenging lighting conditions.

Because event streams differ strongly from conventional image sequences, dedicated representations are typically required before applying standard computer vision or learning-based methods. Common representations include time surfaces, event frames, and voxel grids, each trading off temporal precision, spatial structure, and compatibility with existing algorithms \cite{gallego2022survey}. In recent years, event-based vision has been applied to a wide range of tasks, including feature tracking, optical flow estimation, visual odometry, SLAM, and 3D reconstruction \cite{gallego2022survey}. The development and evaluation of event-based visual odometry methods have also been supported by dedicated datasets and simulators \cite{mueggler2017dataset}. For visual odometry in particular, event cameras are promising because the motion-dependent signal generation directly captures scene changes that are highly informative for pose tracking.

\subsection{Event-Based Visual Odometry}
Visual odometry estimates the motion of a camera from visual measurements over time. In the event-based setting, this task is especially challenging because the measurements are sparse, asynchronous, and highly dependent on scene texture and motion. Early work therefore often relied on geometric formulations and, in many cases, on additional assumptions or sensor modalities. A landmark event-only method is EVO, which demonstrated real-time 6-DoF camera tracking and semi-dense mapping from a single event camera in natural scenes \cite{rebecq2017evo}. However, purely monocular event-only visual odometry remains difficult due to sensitivity to initialization, scene structure, and the sparsity of event data.

To improve robustness, several works incorporated additional sensing modalities. Rebecq \textit{et al.} proposed a real-time visual-inertial odometry pipeline for event cameras based on keyframe optimization \cite{rebecq2017vio}. Ultimate SLAM combined events, standard images, and inertial measurements for improved robustness in HDR and high-speed scenarios \cite{vidal2018ultimate}. Stereo-based approaches such as ESVO exploited two synchronized event cameras to estimate motion and semi-dense depth more reliably \cite{zhou2021esvo}. Further hybrid systems, including Event-Aided Direct Sparse Odometry (EDS), PL-EVIO, and ESVIO, used frame-event fusion, point and line features, or stereo-inertial integration to improve pose estimation accuracy and stability \cite{hidalgo2022eds,guan2022plevio,chen2023esvio}. Although these methods are effective, they generally require more complex hardware and calibration than monocular event-only systems.

\subsection{From DPVO to DEVO}
In parallel with event-based research, learning-based visual odometry in the frame domain advanced significantly. A notable example is Deep Patch Visual Odometry (DPVO), which introduced a sparse patch-based formulation for monocular VO \cite{teed2023dpvo}. Instead of estimating dense flow across full images, DPVO tracks a sparse set of image patches over time, refines correspondences through a recurrent update operator, and jointly optimizes pose and structure using differentiable bundle adjustment \cite{teed2023dpvo}. This sparse formulation achieves strong accuracy while remaining computationally efficient.

Deep Event Visual Odometry (DEVO) transfers this sparse patch-based idea to the event domain. Since event data are naturally sparse and often leave large parts of the image plane empty, DEVO introduces a learned patch selection mechanism that predicts a score map for identifying informative locations for tracking. The selected event patches are then tracked over time, their correspondences are iteratively refined, and camera poses together with sparse depth variables are optimized using differentiable bundle adjustment. The original DEVO paper shows that this design leads to strong monocular event-only visual odometry performance on several real-world benchmarks and substantially improves over prior event-only approaches \cite{klenk2024devo}. This makes DEVO a particularly suitable baseline for our project.

\subsection{Sparse 3D Structure and Point-Cloud Reconstruction}
Although DEVO is designed primarily as a visual odometry system, its optimization process already maintains sparse geometric structure internally through tracked patches and associated depth estimates \cite{klenk2024devo}. This observation is central to our work: if these latent 3D estimates are exposed and transformed into an explicit output format, the odometry pipeline can also provide a sparse point-cloud representation of the scene.

This idea is related to event-based 3D reconstruction methods, but it is important to distinguish our approach from dedicated reconstruction algorithms. For example, EMVS formulates event-based multi-view stereo as a separate 3D reconstruction problem and reconstructs scene structure by combining events from multiple views using known camera poses \cite{rebecq2018emvs}. In contrast, our work does not introduce a new dense or semi-dense reconstruction method. Instead, we reuse the sparse 3D information already estimated inside DEVO and extend the framework with an export and post-processing pipeline. Therefore, the resulting output is best understood as a sparse point cloud derived from DEVO's internal representation rather than as a fully independent reconstruction module.

\section{Method}
This project extends Deep Event Visual Odometry (DEVO) with a mechanism for exporting the sparse 3D scene representation that already exists inside the original odometry pipeline. Our goal is not to replace the core visual odometry formulation of DEVO, but to expose its internal geometric estimates in a form that can be reused for visualization and downstream processing. The resulting method can therefore be understood as a sparse point-cloud extraction pipeline built on top of event-based visual odometry.

\subsection{Overview of the Original DEVO Pipeline}
DEVO formulates monocular event-only visual odometry as a sparse patch-based tracking problem. The input to the system is a sequence of event voxel grids constructed from the raw asynchronous event stream. These voxel grids preserve temporal information by discretizing events into multiple bins and are compatible with convolutional neural networks \cite{klenk2024devo}. Instead of performing dense processing over the full image plane, DEVO tracks only a selected set of event patches over time.

A key component of DEVO is its learned patch selection strategy. Since event data are sparse and often contain large uninformative regions, random or gradient-based patch selection is not sufficiently robust. DEVO therefore predicts a score map that highlights informative regions for optical flow estimation and pose tracking \cite{klenk2024devo}. Based on these selected patches, the system estimates correspondences across multiple event voxel grids. A recurrent update operator iteratively refines the patch motion, while differentiable bundle adjustment jointly optimizes camera poses and sparse depth variables associated with the tracked patches \cite{klenk2024devo,teed2023dpvo}.

As a result, DEVO outputs accurate camera trajectories while internally maintaining a sparse representation of scene structure. This latent structure is encoded through the tracked patch locations, their estimated inverse depths, and the optimized camera poses over time \cite{klenk2024devo}. In the original formulation, these quantities primarily serve the odometry process itself and are not the final output of the system.

\subsection{Motivation for Sparse Point-Cloud Export}
Although trajectory estimation is the primary objective of visual odometry, many practical applications also benefit from explicit spatial scene output. In robotics, an estimated point cloud can support qualitative inspection of the reconstruction, debugging of pose failures, and further geometric processing. In our project, we therefore extend DEVO to make its sparse 3D information accessible outside the internal optimization loop.

The central observation behind our method is that DEVO already estimates the ingredients required for sparse 3D reconstruction: camera poses and patch depths. If these variables are exported during or after inference, each valid tracked patch can be back-projected into 3D space and represented as a scene point. Consequently, the task is not to learn a new reconstruction model, but to expose and transform the existing sparse geometry into an explicit point-cloud representation.

\subsection{Extension of DEVO for Point-Cloud Extraction}
Our extension preserves the original DEVO inference pipeline and adds a data export layer on top of it. During inference, the odometry system still processes event voxel grids, selects informative patches, tracks them over time, and updates camera poses and depths through differentiable bundle adjustment. The main modification is that the relevant intermediate observables are made available to the user instead of remaining internal to the pipeline.

More specifically, the extended system exposes the optimized camera poses together with the sparse geometric state associated with the tracked patches. For each selected patch, the system provides its image-space location and depth-related information. Combined with the corresponding camera pose and camera calibration, these quantities can be converted into 3D coordinates in the camera or world frame. Repeating this process over the sequence yields a sparse set of reconstructed 3D points, which can then be aggregated into a point cloud.

This extension does not alter the optimization objective of DEVO and does not introduce a separate reconstruction loss. The point cloud is therefore derived from the same sparse structure that DEVO already uses internally for visual odometry. As a consequence, the quality and density of the exported point cloud are directly tied to the quality of patch selection, correspondence estimation, and depth optimization in the original DEVO framework.

\subsection{3D Point Generation}
The conversion from sparse patch representation to point-cloud data follows the standard back-projection principle. Let a tracked patch correspond to image coordinates $(u,v)$ and let its estimated depth be $z$. Using the intrinsic camera matrix
\begin{equation}
K =
\begin{bmatrix}
f_x & 0 & c_x\\
0 & f_y & c_y\\
0 & 0 & 1
\end{bmatrix},
\end{equation}
the corresponding 3D point in camera coordinates can be obtained as
\begin{equation}
\mathbf{p}_c = z K^{-1}
\begin{bmatrix}
u\\
v\\
1
\end{bmatrix}.
\end{equation}
Given the estimated camera pose, the point can then be transformed into a common world coordinate frame. By applying this back-projection to all valid sparse patches across the processed sequence, we obtain a set of 3D points that forms the exported sparse point cloud.

Because DEVO operates on a sparse subset of image locations, the resulting cloud is inherently sparse. It does not represent a dense surface reconstruction, but rather a collection of 3D points associated with the most informative tracked event patches. This is consistent with the original design of DEVO, which prioritizes robust sparse tracking over dense scene estimation \cite{klenk2024devo}.

\subsection{Post-Processing and Export Representation}
To make the exported data practically usable, our method includes a post-processing stage after point extraction. First, the generated point data are collected and stored in an intermediate format suitable for further transformation. Next, they are converted into a standard point-cloud representation that can be visualized with common 3D tools. Finally, optional cleanup steps are applied in order to remove outliers, reduce noise, and improve the interpretability of the result.

These post-processing operations do not change the underlying odometry estimates, but they significantly improve the usability of the exported cloud. This separation between odometry and post-processing is important: the visual odometry system remains responsible for estimating motion and sparse geometry, while the export pipeline transforms this information into a form suitable for inspection and downstream applications.

\subsection{Relation to Event-Based Reconstruction Methods}
Our method is related to event-based 3D reconstruction, but it differs in scope and objective from dedicated reconstruction approaches such as EMVS \cite{rebecq2018emvs}. EMVS explicitly formulates multi-view stereo reconstruction from events and known poses, whereas our approach reuses the sparse scene structure already maintained by DEVO. In other words, our contribution is not a new dense reconstruction method, but an extension that exposes DEVO's latent geometry as an explicit output.

For this reason, the result should be interpreted as a sparse point cloud derived from event-based visual odometry rather than as a complete scene model. Nevertheless, this sparse output is valuable in practice because it bridges the gap between trajectory-only estimation and geometric scene representation. It also creates a useful interface for further processing stages, including visualization, filtering, or integration with additional reconstruction methods.\\

In summary, the proposed method augments DEVO with sparse point-cloud export while preserving the original event-based odometry pipeline. The system continues to estimate camera poses and sparse depth variables using learned patch selection, recurrent refinement, and differentiable bundle adjustment. Our extension exposes these internal variables, converts them into 3D point coordinates, and prepares them for visualization and further processing. The method therefore turns DEVO from a trajectory-focused event-based VO system into a framework that can additionally provide sparse geometric scene output.

\section{Implementation and Pipeline}
The practical implementation of our project is organized as an extension of the original DEVO codebase rather than as a separate reconstruction framework. The repository preserves the original project structure, including the main \texttt{devo}, \texttt{config}, \texttt{evals}, \texttt{scripts}, and \texttt{utils} directories, and adds documentation and utilities specifically for point-cloud export.

\subsection{Code-Level Extension of DEVO}
The central implementation change is an opt-in export mechanism for internal DEVO observables. In our modified version, the helper entry points \texttt{run\_rgb}, \texttt{run\_voxel\_norm\_seq}, and \texttt{run\_voxel} were extended with a \texttt{return\_observables=True} flag. When enabled, these functions return not only poses and timestamps, but also the current sparse point cloud and depth estimates. This modification is backward compatible, since the original behavior remains unchanged when the flag is omitted.

This implementation choice is important because it keeps the odometry pipeline intact. Rather than rewriting DEVO's optimization logic, the project exposes quantities that DEVO already computes internally. DEVO stores event patches and sparse geometric state during the update process, and our extension makes these observables accessible outside the original inference loop.

\subsection{Export Script}
The main export functionality is implemented in \texttt{scripts/export\_pointcloud.py}. This script supports different datasets through a command-line interface and selects dataset-specific iterators and geometry settings for FPV, RPG, and VECtor-style data. The script also exposes parameters for output naming, side selection, stride, timestamps, optional visualization, and additional export modes such as per-frame cloud saving and frame-data export. In addition, it supports conversion to \texttt{.ply} and optional cleanup of the generated cloud.

From an implementation perspective, this script acts as the bridge between DEVO inference and explicit geometric output. It prepares the appropriate iterator, loads calibration when available, runs the export routine, and writes the resulting sparse point cloud into a format that can be used outside the original odometry process. This makes the export step reproducible and configurable from the command line.

\subsection{Pipeline}
To simplify the full workflow, the repository provides \texttt{scripts/start\_pipline.py} as the main orchestration script. This entry point is designed to run the complete point-cloud pipeline for a given input folder. Its command-line interface includes options for the input dataset path, model weights, configuration file, output names, dataset type, export side, cleanup algorithm, and optional downstream stages.

The orchestration logic supports a multi-stage workflow. First, it checks whether preprocessing outputs required for VECtor-style export are available. Then it resolves the dataset path layout and launches the export script as a subprocess. After export, the pipeline can convert the generated NumPy cloud into a point-cloud file format and optionally perform cleanup. The script additionally exposes options related to EMVS-oriented export and downstream execution.

To simplify deployment of the proposed system, we additionally prepared a Docker container for the full pipeline \cite{safdari2026devoDocker}. This was particularly useful because the workflow includes several setup stages, software dependencies, dataset-specific configurations, and optional export and cleanup steps, which are time-consuming and difficult to reproduce manually. The containerized environment provides a consistent execution setup for preprocessing, inference, point-cloud export, and post-processing. As a result, it improves portability, reduces configuration errors, and makes the overall system easier to use and evaluate across different machines.

\subsection{Data Products}
The modified implementation produces several useful outputs. At the most basic level, the fork can return estimated poses, timestamps, a sparse point cloud, and patch depths directly from the DEVO helper functions. These data are returned as NumPy arrays and can be exported to formats such as \texttt{.ply} or \texttt{.npy}. Beyond this, the export script also supports saving per-frame clouds, bundling frame-wise metadata into \texttt{.npz} files, and saving accumulated viewer-style point data.

These outputs are useful because they separate the geometric information from the online odometry loop. As a result, the same run can be used both for pose estimation and for subsequent inspection of the recovered sparse 3D structure. This is especially valuable in a project setting, where qualitative analysis of the generated geometry is as important as the trajectory itself.

\subsection{Post-Processing Support}
The implementation includes explicit support for post-processing after export. The command-line interface of \texttt{export\_pointcloud.py} exposes cleanup options such as statistical outlier removal and radius outlier removal, together with their associated hyperparameters. The pipeline wrapper forwards cleanup options as part of the full workflow, allowing raw sparse point clouds to be converted and filtered in a single execution path.

This design choice is practically important. Since the exported geometry is sparse and derived from odometry state rather than dense reconstruction, the raw point cloud can contain isolated points and local noise. Cleanup therefore does not change the underlying DEVO estimates, but improves readability and makes the final output more useful for visualization and later processing.

\begin{figure}[!t]
    \centering
    \includegraphics[height=20cm, keepaspectratio]{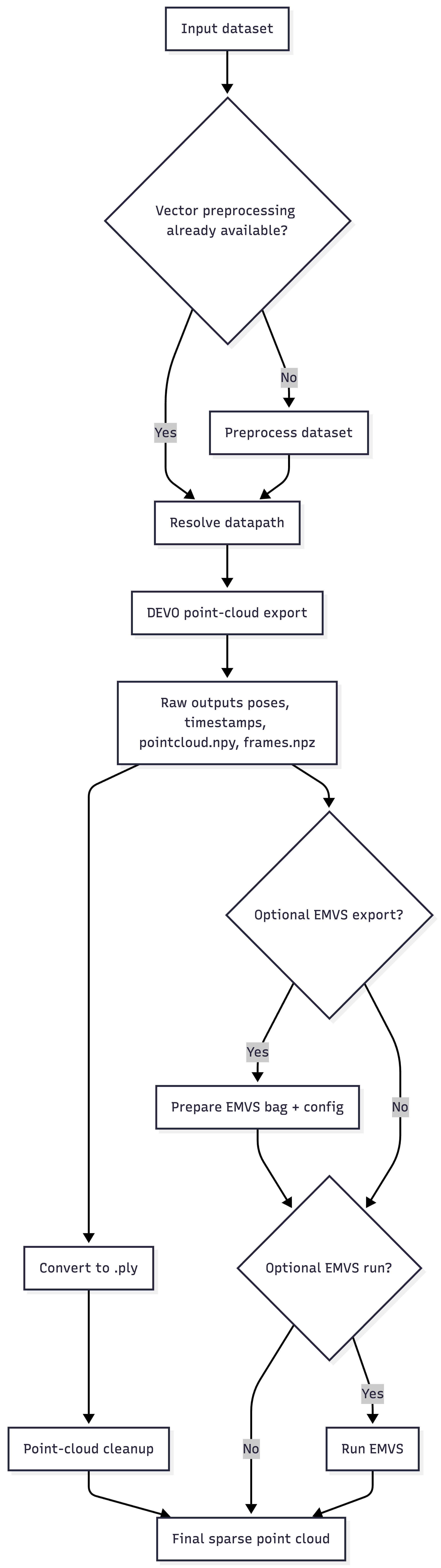}
    \caption{Workflow of the extended DEVO point-cloud pipeline}
\end{figure}

Overall, the implementation can be summarized as follows. First, event data are prepared and passed to the DEVO-based inference pipeline. Second, the modified helper functions expose sparse geometric observables at the end of a run. Third, \texttt{export\_pointcloud.py} writes these observables into explicit point-cloud outputs. Finally, \texttt{start\_pipline.py} automates preprocessing, export, conversion, cleanup, and optional downstream integration. The result is a practical end-to-end workflow that extends DEVO from pure event-based visual odometry toward sparse geometric scene export.

\section{Experiments and Results}
We evaluated the proposed point-cloud export pipeline on the \textit{BOARD SLOW} event-camera sequence. The scene contains a planar calibration board with high-contrast geometric patterns, including diamond, square, circular, and symbol-like regions. This sequence is well suited for validating event-based sparse reconstruction because event cameras mainly respond to brightness changes; therefore, the strongest reconstructed structures are expected along the printed shape boundaries on the board.

The goal of the experiment was not to obtain a dense surface reconstruction. Instead, we aimed to determine whether the sparse 3D structure exported from the modified DEVO pipeline is geometrically meaningful and whether it is consistent with an independent event-based reconstruction. For this purpose, we compared the exported DEVO SLAM point cloud with EMVS reconstructions generated from the same event sequence using both DEVO poses and ground-truth poses. The validation scripts and figure-generation code for this experiment are provided in the accompanying Jupyter notebook~\cite{safdari2026notebook}.

\subsection{Experimental Setup}
Three point-cloud representations were evaluated. The first was the sparse point cloud exported directly from the modified DEVO pipeline, referred to as the \textit{DEVO SLAM cloud}. The second was an EMVS reconstruction generated using DEVO poses. The third was an EMVS reconstruction generated using ground-truth poses. The EMVS reconstructions were used as reference reconstructions because EMVS explicitly reconstructs semi-dense scene structure from events and known camera poses.

The BOARD SLOW sequence contained $985$ DEVO poses, $4171$ ground-truth poses, and $1043$ image timestamps. The exported DEVO SLAM cloud contained $7717$ points. The EMVS reconstruction using ground-truth poses contained $4444$ points, while the EMVS reconstruction using DEVO poses contained $2993$ points. Thus, the ground-truth-pose EMVS reconstruction produced approximately $1.48\times$ more 3D points than the DEVO-pose EMVS reconstruction.

A scale mismatch was observed between the DEVO trajectory and the ground-truth trajectory. The DEVO trajectory was approximately $1.4184\times$ larger than the ground-truth metric scale. Equivalently, the ground-truth scale corresponded to approximately $0.705$ times the DEVO scale. To separate scale effects from pose-source effects, four EMVS variants were evaluated: ground-truth poses at true metric scale, scaled ground-truth poses, original DEVO poses, and de-scaled DEVO poses. The depth ranges were adjusted consistently with the corresponding pose scale.

\begin{table}[H]
\centering
\caption{Input data and reconstructed point-cloud sizes for the BOARD SLOW experiment.}
\label{tab:board_slow_data}
\begin{tabular}{|l|c|}
\hline
\textbf{Quantity} & \textbf{Value} \\
\hline
DEVO poses & 985 \\
Ground-truth poses & 4171 \\
Image timestamps & 1043 \\
DEVO SLAM cloud points & 7717 \\
EMVS with DEVO poses & 2993 points \\
EMVS with ground-truth poses & 4444 points \\
GT/DEVO scale factor & 0.705 \\
DEVO/GT scale factor & 1.4184 \\
\hline
\end{tabular}
\end{table}

\subsection{Scale Isolation Experiment}
An initial hypothesis was that DEVO's approximate $1.42\times$ scale overestimation might explain the apparent quality difference between the EMVS reconstructions. However, the scale isolation experiment showed that this hypothesis is not supported. When both the pose baseline and the EMVS depth range were scaled consistently, the reconstruction quality did not change in a way that would explain the observed difference.

This result is expected because EMVS mainly depends on angular multi-view geometry. If the camera baseline and the scene depth are scaled by the same factor, the event ray geometry and disparity structure remain largely unchanged. Therefore, uniform scale alone does not explain the reconstruction-quality difference between the DEVO-pose and ground-truth-pose EMVS outputs.

The corrected interpretation is that the ground-truth-pose EMVS reconstruction is better mainly because of pose density. The ground-truth trajectory provides $4171$ poses over the sequence, while DEVO provides $985$ poses. This corresponds to approximately $4.2\times$ denser pose sampling. 

\begin{figure}[H]
    \centering
    \includegraphics[width=\linewidth]{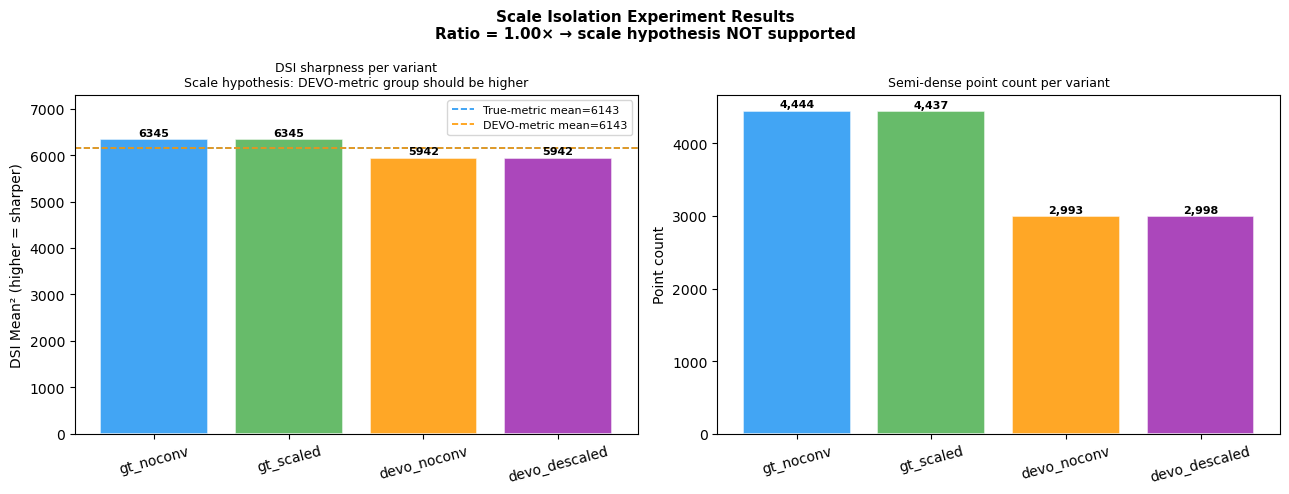}
    \caption{Scale isolation experiment on the BOARD SLOW sequence. Scaling the poses and depth range consistently does not improve reconstruction sharpness or point count, showing that the scale hypothesis is not supported. The main difference is caused by the pose source and pose density.}
    \label{fig:scale_isolation}
\end{figure}

The denser ground-truth trajectory gives EMVS more independent viewpoints per event, resulting in sharper DSI peaks and more confident reconstructed points.

\begin{table}[H]
\centering
\caption{Summary of the scale and pose-density findings.}
\label{tab:scale_pose_density}
\begin{tabular}{|l|c|c|}
\hline
\textbf{Pose source} & \textbf{Number of poses} & \textbf{EMVS points} \\
\hline
Ground truth & 4171 & 4444 \\
DEVO & 985 & 2993 \\
\hline
\end{tabular}
\end{table}

\subsection{Point-Cloud Agreement Between EMVS and DEVO SLAM}
To evaluate whether the exported DEVO point cloud represents the same physical structure as the event-based reconstruction, the EMVS reconstruction generated with DEVO poses was transformed into the DEVO world frame and compared with the exported DEVO SLAM cloud. We used nearest-neighbor distance, F-score, precision, recall, and ICP alignment.

The one-directional Chamfer distance from the EMVS cloud to the DEVO SLAM cloud was $9.1$ mm before ICP and $9.4$ mm after ICP. The very small change after ICP indicates that the two clouds were already well aligned before rigid registration. Therefore, the remaining differences are not mainly caused by a global rigid-frame error.

At a $5$ cm threshold, the F-score was $0.479$ before ICP and $0.491$ after ICP. The precision was $98.4\%$, while the recall was $31.7\%$. The high precision means that almost all EMVS points lie close to the exported DEVO SLAM cloud. The lower recall is expected because the two clouds have different coverage properties: EMVS reconstructs a more localized semi-dense structure from the selected event window, whereas the DEVO SLAM cloud accumulates sparse points over the full trajectory.

\begin{table}[H]
\centering
\caption{Comparison between EMVS using DEVO poses and the exported DEVO SLAM cloud.}
\label{tab:emvs_devo_slam}
\begin{tabular}{|l|c|}
\hline
\textbf{Metric} & \textbf{Result} \\
\hline
Chamfer EMVS$\rightarrow$SLAM before ICP & 9.1 mm \\
Chamfer EMVS$\rightarrow$SLAM after ICP & 9.4 mm \\
F-score @ 5 cm before ICP & 0.479 \\
F-score @ 5 cm after ICP & 0.491 \\
Precision @ 5 cm & 98.4\% \\
Recall @ 5 cm & 31.7\% \\
\hline
\end{tabular}
\end{table}

The 3-D overlay of the two clouds showed the same board-level structure in the DEVO world frame, but the DEVO SLAM cloud covered a larger accumulated area and contained more isolated outliers. Therefore, the numerical agreement should be interpreted as local consistency rather than full one-to-one cloud overlap.

This supports the interpretation that the exported DEVO cloud is geometrically valid locally, but should not be interpreted as a complete reconstruction of the whole board.

\subsection{Planarity Validation}
Since the observed object is a planar board, we evaluated planarity by fitting a best-fit plane to each point cloud and measuring the point-to-plane residuals. The EMVS reconstruction using DEVO poses achieved a plane-fitting RMSE of $13.5$ mm. The exported DEVO SLAM cloud had a higher RMSE of $93.2$ mm.

\begin{table}[H]
\centering
\caption{Planarity evaluation using residuals to a best-fit plane.}
\label{tab:planarity_results}
\begin{tabular}{|l|c|}
\hline
\textbf{Point cloud} & \textbf{Plane RMSE} \\
\hline
EMVS using DEVO poses & 13.5 mm \\
Exported DEVO SLAM cloud & 93.2 mm \\
\hline
\end{tabular}
\end{table}

The lower EMVS residual is expected because EMVS concentrates reconstructed points on event-supported surfaces in a selected view. In contrast, the DEVO SLAM cloud is an accumulated sparse odometry output. It contains points collected over the full camera trajectory and is therefore more affected by pose drift, tracking uncertainty, and isolated outliers. Nevertheless, both reconstructions recover the same dominant board plane, which confirms that the exported DEVO structure is geometrically meaningful.

\subsection{Comparison with Ground-Truth-Pose EMVS}
We also compared the ground-truth-pose EMVS reconstruction with the exported DEVO SLAM cloud. Because the ground-truth and DEVO trajectories are expressed in different coordinate frames and scales, the ground-truth-based EMVS cloud was first scaled to the DEVO metric using the observed scale factor and then approximately transformed into the DEVO world frame.

After scale and frame normalization, the ground-truth-pose EMVS cloud achieved an F-score of $0.465$ at a $5$ cm threshold, with precision of $91.3\%$ and recall of $31.2\%$. These values are close to the EMVS-DEVO comparison. This indicates that the exported DEVO SLAM cloud is consistent not only with EMVS using DEVO poses, but also with an independently generated EMVS reconstruction based on ground-truth poses.

\begin{table}[H]
\centering
\caption{Comparison between ground-truth-pose EMVS, approximately transformed into the DEVO world frame, and the exported DEVO SLAM cloud.}
\label{tab:gt_emvs_devo_comparison}
\begin{tabular}{|c|c|c|c|}
\hline
\textbf{Threshold} & \textbf{F-score} & \textbf{Precision} & \textbf{Recall} \\
\hline
1 cm & 0.142 & 0.307 & 0.092 \\
2 cm & 0.289 & 0.666 & 0.185 \\
5 cm & 0.465 & 0.913 & 0.312 \\
10 cm & 0.526 & 0.969 & 0.361 \\
\hline
\end{tabular}
\end{table}

\begin{figure}[!t]
    \centering
    \includegraphics[width=0.82\linewidth]{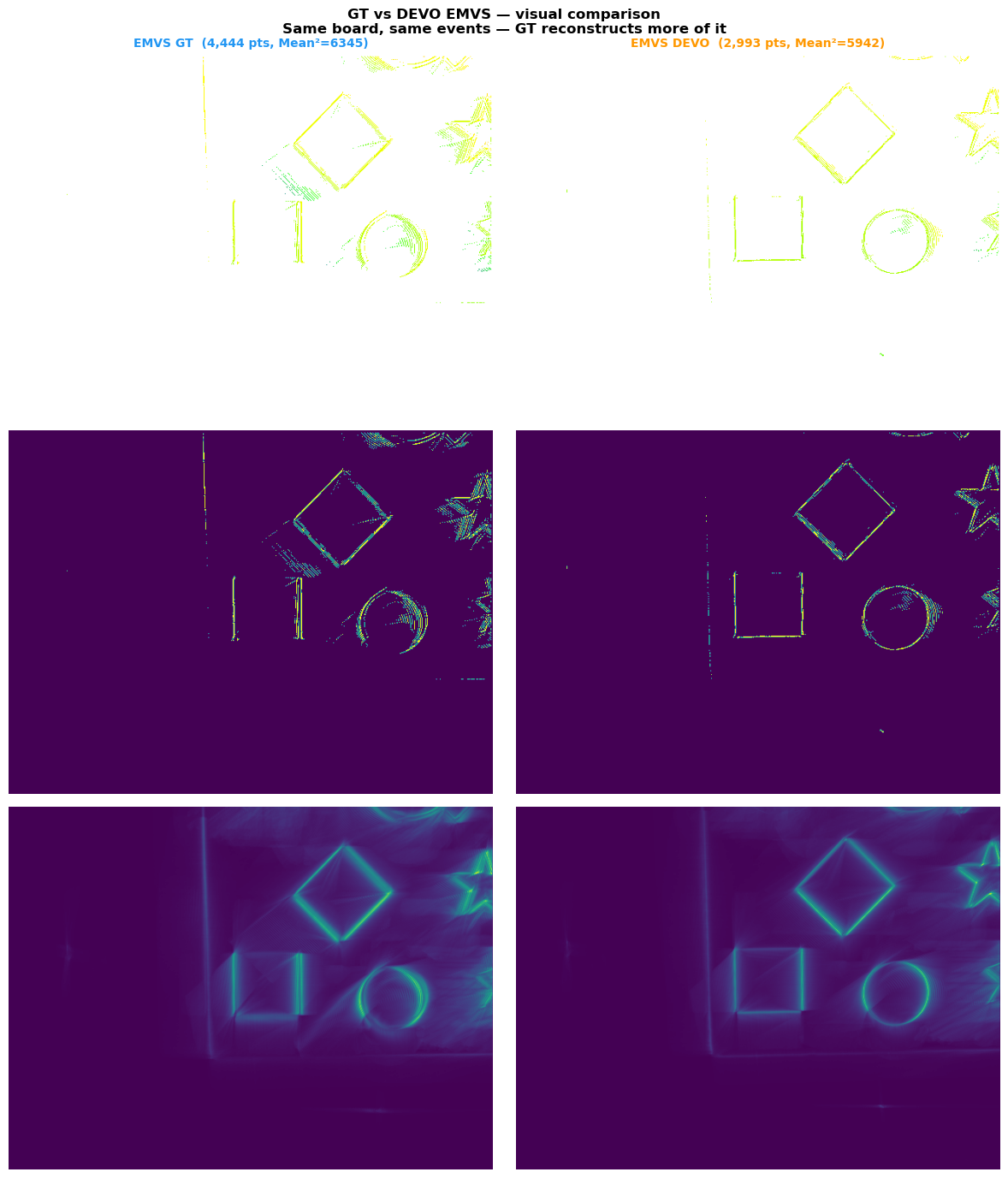}
    \caption{Visual comparison between EMVS reconstructions using ground-truth poses and DEVO poses. Both reconstructions recover the same board shape boundaries, but the ground-truth-pose reconstruction produces more complete event-supported structure.}
    \label{fig:gt_devo_emvs_visual}
\end{figure}

The visual comparison in Fig.~\ref{fig:gt_devo_emvs_visual} confirms this interpretation: both pose sources reconstruct the same board-edge structures, but the ground-truth-pose reconstruction preserves more complete edge support.

The slightly lower precision compared with the DEVO-pose EMVS comparison should not be interpreted as worse geometric accuracy of the ground-truth reconstruction. It is partly caused by the approximate transformation from the ground-truth frame into the DEVO world frame. In the reference-view camera frame, both EMVS variants reconstruct the same board geometry, including the diamond, square, and circular shape boundaries.

\subsection{Image-Space Shape Validation}
Qualitative inspection confirms the quantitative findings. The reconstructed points concentrate around the printed shape boundaries of the calibration board. This behavior is consistent with the event-camera sensing principle: events are triggered primarily by brightness changes, and therefore the edges of the printed patterns generate the strongest reconstruction support.

\begin{figure}[H]
    \centering
    \includegraphics[width=\linewidth]{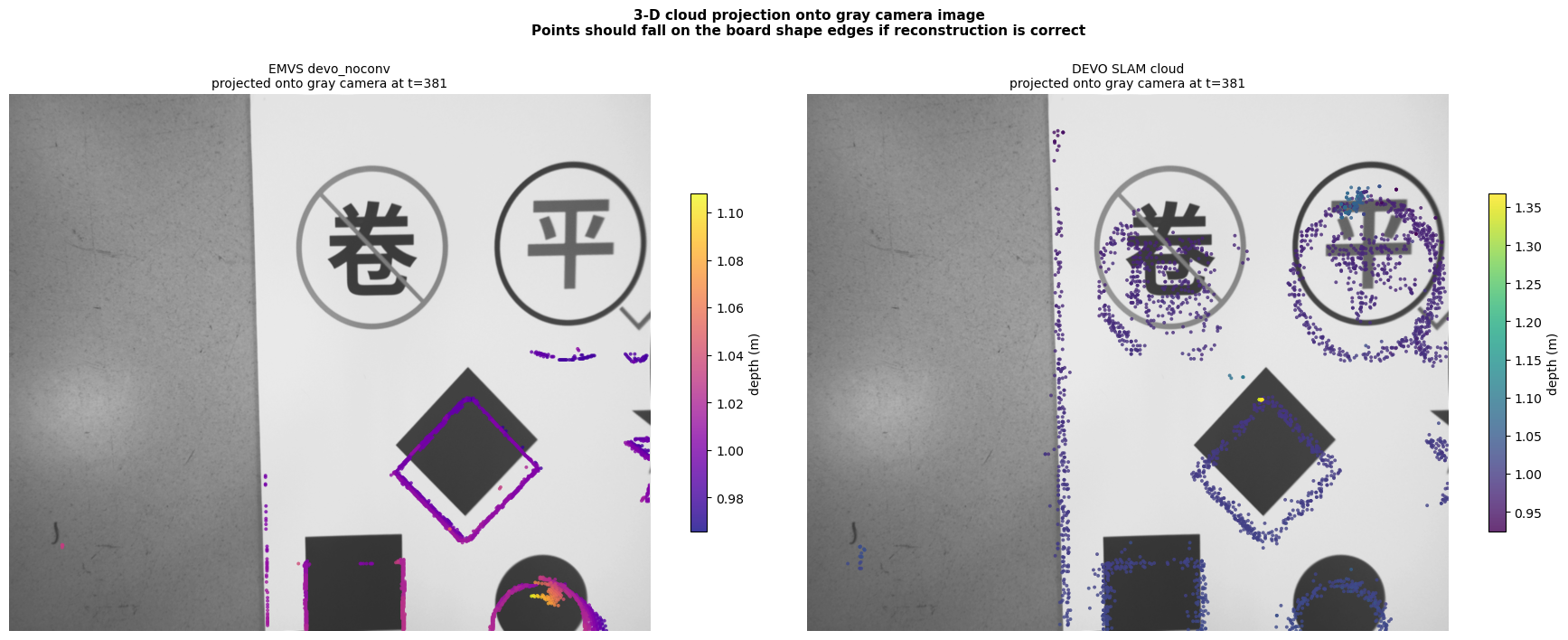}
    \caption{Projection of reconstructed 3D points onto a regular gray-camera frame. The points align with visible board shape boundaries such as the diamond, circle, square, and printed symbols, confirming that the exported point clouds correspond to real scene geometry.}
    \label{fig:cloud_projection_gray}
\end{figure}

To verify this directly, the reconstructed 3D points were projected back onto a regular gray-camera frame, as shown in Fig.~\ref{fig:cloud_projection_gray}.

Both EMVS and DEVO SLAM recover the dominant planar board structure. The EMVS reconstruction is more compact and has lower plane residuals, while the exported DEVO SLAM cloud is sparser and noisier because it directly reflects the accumulated sparse odometry state. However, the main board geometry is still visible in the exported DEVO cloud, which confirms that the proposed export mechanism exposes meaningful 3D structure rather than arbitrary internal optimization variables.
 
\section{Discussion}
The experiments show that the proposed extension successfully turns DEVO from a trajectory-only event-based VO pipeline into a system that also provides usable sparse scene geometry.

The BOARD SLOW experiment provides three concrete findings. First, the scale-isolation experiment shows that the quality difference between ground-truth-pose EMVS and DEVO-pose EMVS is not primarily caused by the global DEVO scale factor. Scaling the trajectory and depth range consistently did not improve the DEVO-pose reconstruction. Second, the main factor is pose density: the ground-truth trajectory provides approximately $4.2\times$ more poses than the DEVO trajectory, which gives EMVS denser multi-view support and results in a more complete semi-dense reconstruction. Third, the exported DEVO SLAM cloud is locally consistent with EMVS, as shown by the high $98.4\%$ precision at a $5$ cm threshold, but it remains noisier and less planar than the EMVS output because it accumulates sparse odometry structure over the full trajectory.

At the same time, the experiments also highlight the limitations of the approach. The exported point cloud is inherently sparse because DEVO only tracks a selected subset of informative patches rather than performing dense depth estimation over the entire image. As a result, the reconstruction does not form a complete surface model of the scene. It is more accurately interpreted as a sparse structural representation that is useful for visualization, inspection, and downstream processing, but not as a dense reconstruction comparable to dedicated multi-view stereo methods such as EMVS \cite{rebecq2018emvs}.

Another limitation is that the quality of the exported geometry depends directly on the stability of the underlying odometry process. Errors in patch tracking, depth estimation, or pose optimization can appear as noisy or isolated 3D points in the final cloud. For this reason, post-processing plays an important practical role in improving the readability of the exported result, even though it does not change the underlying DEVO estimates.

Nevertheless, the project demonstrates a useful intermediate direction between pure visual odometry and full 3D reconstruction. By exposing DEVO's latent scene structure, the system becomes more useful for robotics workflows that require not only motion estimation but also a lightweight geometric representation of the environment. Future work could investigate denser fusion across time, improved filtering, or tighter integration with dedicated reconstruction methods in order to increase completeness and robustness.

\section{Conclusion}
In this report, we presented an extension of Deep Event Visual Odometry (DEVO) for sparse point-cloud export. DEVO already estimates sparse scene structure internally through tracked event patches, depth variables, and optimized camera poses. Our contribution was to expose this latent geometric information and transform it into an explicit point-cloud representation that can be exported, visualized, and post-processed.

The proposed extension preserves the original event-based visual odometry pipeline while increasing its practical usefulness for scene inspection and downstream geometric processing. Experimental observations showed that the modified system can successfully produce sparse point clouds from event-based sequences, while also confirming the expected limitations in density, completeness, and sensitivity to outliers.

On the BOARD SLOW sequence, the exported DEVO cloud aligned with event-supported EMVS structure and preserved the dominant planar board geometry, while the scale-isolation experiment showed that pose density, rather than global scale error, was the main reason for the difference between ground-truth-pose and DEVO-pose EMVS reconstructions.

Overall, the project demonstrates that DEVO can be extended from a trajectory-focused odometry framework toward a more versatile system that additionally provides sparse 3D scene output. This creates a useful basis for future work on improved filtering, temporal fusion, and integration with more advanced event-based reconstruction methods.

\appendices
\section{Reproducibility Material}
The quantitative and qualitative validation results reported in Section~V were produced using the accompanying Jupyter notebook
\texttt{board\_slow\_shape\_validation.ipynb}. The notebook loads the BOARD SLOW calibration files, DEVO poses, ground-truth poses, EMVS point clouds, and the exported DEVO SLAM cloud. It then performs the scale-isolation experiment, point-cloud projection checks, ICP alignment, Chamfer/F-score evaluation, and board-planarity analysis.

The notebook also generates the figures used in the experimental evaluation, including the scale-isolation plot, EMVS ground-truth versus DEVO-pose comparison, and the image-space projection of reconstructed 3D points onto the regular gray-camera frame.

The full notebook is provided as supplementary material together with this report and is available in the project repository~\cite{safdari2026notebook}.

\section*{Acknowledgement}
AI-based tools, including GPT models and integrated AI features in Overleaf, were used during the preparation of this report to support literature search and filtering. In addition, these tools were used to improve sentence formulation and overall wording in order to communicate the main ideas of the project more clearly and effectively. All sources were reviewed by the authors, and the authors are responsible for the final selection of references, the technical interpretation of the cited work, and the final written content of the report.

\bibliographystyle{IEEEtran}
\bibliography{EventBasedRobotVisionPROJECT}

\end{document}